%% file: main_arXiv.tex
\documentclass[10pt,twocolumn,letterpaper]{article}

\usepackage{cvpr}
\usepackage{times}
\usepackage{epsfig}
\usepackage{graphicx}
\usepackage{amsmath}
\usepackage{amssymb}
\usepackage{tablefootnote}
\usepackage{tabularx}
\usepackage{amsmath,epsfig,bm,amssymb,subfigure,graphicx,algorithm,algorithmic,color}
\input{mathdef}

\usepackage[pagebackref=true,breaklinks=true,letterpaper=true,colorlinks,bookmarks=false]{hyperref}

\ifcvprfinal\fi

\newcommand{\mname}[1]{Deep Matching Autoencoders}
\newcommand{\mabbr}[1]{DMAE}
\newcommand{\tname}[1]{unsupervised classifier learning}
\newcommand{\keypoint}[1]{\vspace{0.1cm}\noindent\textbf{#1}\quad}

\newlength\mylength
\usepackage[breaklinks=true,bookmarks=false]{hyperref}

\cvprfinalcopy % *** Uncomment this line for the final submission

\def\cvprPaperID{2177} % *** Enter the CVPR Paper ID here

\begin{document}

%%%%%%%%% TITLE
\title{Deep Matching Autoencoders}

\author{Tanmoy Mukherjee \\
University of Edinburgh\\
{\tt\small mukherjee.tanmoy@gmail.com}
\and
Makoto Yamada\\
RIKEN AIP, JST PRESTO\\
{\tt\small makoto.yamada@riken.jp}
\and 
Timothy M. Hospedales\\
University of Edinburgh\\
{\tt\small t.hospedales@ed.ac.uk}
}

\maketitle
%\thispagestyle{empty}

%%%%%%%%% ABSTRACT
\begin{abstract}
Increasingly many real world tasks involve data in multiple modalities or views.  This has motivated the development of many effective algorithms for learning a common latent space to relate multiple domains. However, most existing cross-view learning algorithms assume access to paired data for training. Their applicability is thus limited as the paired data assumption is often violated in practice: many tasks have only a small subset of data available with pairing annotation, or even no paired data at all.

In this paper we introduce \mname~~(\mabbr~), which learn a common latent space and pairing from {\bf unpaired} multi-modal data. Specifically we formulate this as a cross-domain representation learning and object matching problem. We simultaneously optimise parameters of representation learning auto-encoders and the pairing of { unpaired} multi-modal  data. This framework elegantly spans the full regime from fully supervised, semi-supervised, and unsupervised (no paired data) multi-modal learning. We show promising results in image captioning, and on a new task that is uniquely enabled by our methodology: \tname~.
\end{abstract}

%% Plan.
% P1: Multi-modal representation learning problem. Breadth and significance.
% P2: Multi-modal representation learning methods. What are they and what's wrong with them.
% P3. Related methods and why they are not good enough.
% P4. Our method. 

\section{Introduction}

% P1: Multi-modal representation learning problem. Breadth and significance.
Learning representations from multi-modal data is a widely relevant problem setting in many applications of machine learning and pattern recognition. In computer vision it arises in tagging \cite{feng2014correspondenceAutoencoder,gong2013embedding}, cross-view \cite{gong2014reidBook,kan2016multiViewDeep} and cross-modal \cite{ouyang2016heteroFaceSurvey} learning. It is particularly relevant at the border between vision and other modalities, for example audio-visual speech classification \cite{ICML2011Ngiam_399} and generating descriptions of images and videos \cite{Coyne:2001:WAT:383259.383316,6751448,Gupta:2012:CLO:2900728.2900815,Krishnamoorthy:2013:GNV:2891460.2891535,ordonez2011im2text} in the case of audio and text respectively.

%% P2: Multi-modal representation learning methods. What are they and what's wrong with them.
The wide applicability of multi-modal representation learning has motivated the study of numerous cross-modal learning methods including Canonical Correlation Analysis (CCA) \cite{Hardoon:2004:CCA:1119696.1119703,hotelling1936relations} and Kernel CCA \cite{Bach:2003:KIC:944919.944920}. Progress has further accelerated recently with the contribution of large parallel datasets \cite{Lin2014,TACL229}, which have permitted the application of deep multi-modal models such as DeepCCA \cite{pmlr-v28-andrew13} and other two branch deep networks to tasks such as image-caption matching \cite{Wang2017LearningTN} and zero-shot learning \cite{frome2013devise}. Nevertheless a pervasive limitation of all these methods is that they are fully supervised methods in the sense that they require \emph{paired} training data to learn the cross-modal mapping or embedding space. However, in many applications paired data may be relatively sparse compared to unpaired data, in which case semi-supervised cross-modal learning methods would be beneficial to exploit the abundant unpaired data. Moreover, in some cases it may be desirable to learn from pools of data in each modality which are completely unpaired, necessitating unsupervised cross-modal learning. 

% P3. Related methods and why they are not good enough.
In this paper we address the task of cross-modal learning from partially or completely \emph{unpaired} data. There have been only a few prior attempts to address inferring pairings from partially or completely unmatched data. These include Kernelized sorting (KS) \cite{djuric2012convex,Jeb04,quadrianto2009kernelized}, least-square object matching (LSOM) \cite{yamada2015cross,yamada2011cross}, and matching CCA (MCCA) \cite{haghighi2008learning}.  However these existing algorithms are all \emph{shallow} and thus may not perform well on challenging complex data where representation learning is important, such as images and text. We introduce \mname ~~(\mabbr~), which to our knowledge provide the first deep representation learning approach to unpaired cross-modal learning. 

% P4. Our method. 
Our ~\mabbr~ method employs an auto-encoder model for each data view, which are learned by minimizing reconstruction error as usual. We further introduce a latent alignment matrix to model the unknown pairing between views, which we optimize using cross-modal dependency measures  kernel target alignment (KTA) \cite{cristianini2002kernel} and squared-loss mutual information (SMI) \cite{yamada2015cross}. With this framework we simultaneously learn the autoencoding representation and the cross-view pairing. In this way the representation is trained to support cross-view matching. During training the learned representation improves as cross-view matching is progressively disambiguated, and cross-modal items are paired more accurately as the learned representation progressively improves. 

%% P5. Our method (2)

Our proposed framework elegantly spans the spectrum from fully supervised to fully unsupervised cross-modal learning. The \emph{fully supervised} case corresponds to conventional cross-modal learning, where our approach is an alternative to DeepCCA \cite{pmlr-v28-andrew13} or two branch matching nets \cite{Wang2017LearningTN}, except that we use a statistical dependency-based  rather than correlation or ranking-based loss. We show that our approach performs comparably to state of the art alternatives for supervised tasks. More interestingly, our approach is effective for \emph{semi-supervised} learning (only subset of pairings available), and we show that it is able to better exploit unlabeled multi-modal data to improve performance compared to alternatives such as matching CCA \cite{haghighi2008learning}. % and Kernelized Sorting \cite{quadrianto2009kernelized,Jeb04}. 
 Most interestingly, our approach is even effective for \emph{unsupervised} cross-modal learning where no pairings are given. We demonstrate this capability by introducing and solving a novel task termed \tname ~~(UCL). 

{In the UCL task we assume a pool of unlabelled images are given along with a pool of category embeddings (e.g., word-vectors) that describe the images in the pool. However it is \emph{unsupervised} in that no pairings between images and categories are given.  This task corresponds to an application where we have a pool of images and we have some idea of  the classes likely to be represented in those images; but no specific class-image pairings. Based on these inputs alone we can train classifiers to recognise the categories represented in the category embedding pool. Like the classic clustering problem, this task is unsupervised in that there is no supervision/pairing given. However like the conventional supervised learning setting, UCL produces classifiers for specific nameable image categories as an output. This task can be seen as an extreme version of zero-shot learning \cite{lampert2009zeroshot_dat,DBLP:journals/corr/TsaiHS17}, where there is \emph{no} auxiliary set with image + class embedding pairs available to learn an image-category embedding mapping. The image-category mapping must be learned in an entirely unsupervised way.}

\begin{figure}[t]
\begin{center}
\centering
  {\includegraphics[width=0.99\columnwidth]{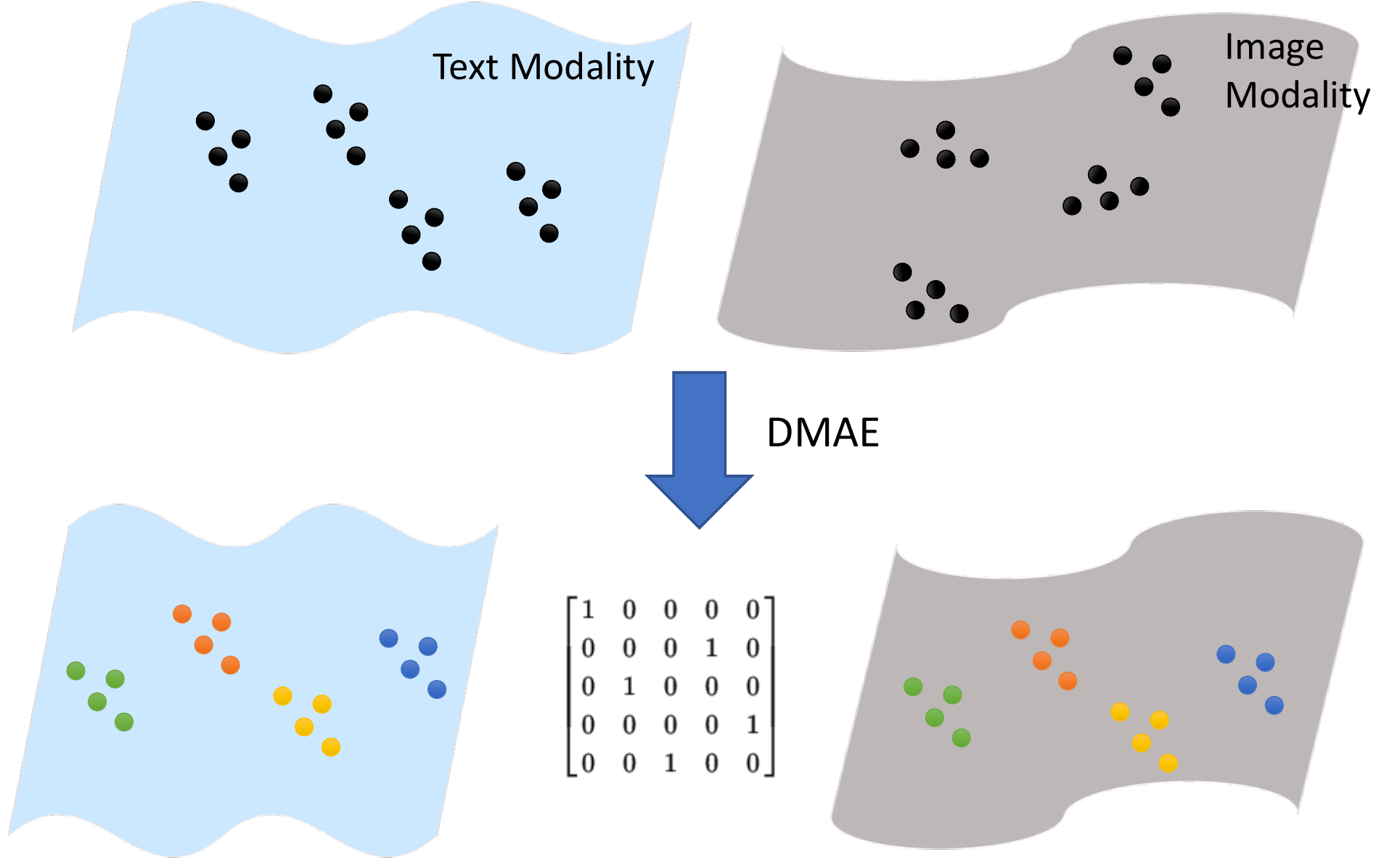}} 
\caption{Multimodal learning from unpaired data problem setting. DMAE inputs a set of unpaired instances in each view and learns: (i) A permutation matrix associating objects across views. (ii) A new representation for each with maximum statistical dependency. }
\label{fig:mdae-arch-1}
\end{center}
\end{figure}

Our contributions can be summarized as follows:

\begin{itemize}
\item We propose \mabbr~, a  cross-view learning and matching framework that elegantly spans supervised, semi-supervised and unsupervised cross-modal learning.
\item Our framework performs comparably to state of the art on the topical image-caption matching benchmark. 
\item \mabbr~~ is effective in leveraging unpaired data in a semi-supervised setting, and learning entirely from unpaired data in an unspervised setting.
\item We introduce and provide a first solution to the the novel problem of unsupervised classifier learning.
\end{itemize}

\section{Related Work}
\label{sec:Related Work}
%In this section, we briefly review some work on multi-modal learning algorithms and its applications. 

\subsection{Multi-modal learning}
Many modern digital events are inherently {multimodal} in nature, i.e a video or image that you favourite is followed with a caption, a tag or comment. In most supervised multi-modal learning setups, it is a privilege to have access to paired data (i.e., $\{(\boldx_i, \boldy_i)\}_{i=1}^n$). For example where $\boldx$ is a vector of image and $\boldy$ is a vector of text. In unsupervised multi-modal learning setup, we can only access to unpaired data $\{\boldx_i\}^n_{i=1}$ and $\{\boldy_j\}^n_{j=1}$. The semi-supervised setup is a mixture of the supervised and unsupervised setup. 

\keypoint{Supervised multi-modal learning} 
The most established supervised multi-modal learning algorithm is  canonical correlation analysis (CCA) \cite{hotelling1936relations}, which learns a linear projection of features in two views such that are maximally correlated in a common latent space. CCA has been studied extensively and has a number of useful properties \cite{Hardoon:2004:CCA:1119696.1119703}. In particular, the optimal linear projection mapping can be obtained by solving an eigenvalue decomposition. It has also been extended to the non-linear case via kernelization (KCCA) \cite{Bach:2003:KIC:944919.944920}.

The huge success of deep neural network (DNN) in computer vision and NLP has inspired many deep multi-modal learning algorithms including DeepCCA \cite{pmlr-v28-andrew13}, multi-modal deep autoencoders (DAEs) \cite{feng2014correspondenceAutoencoder,ICML2011Ngiam_399}, and two branch matching or ranking networks \cite{Wang2017LearningTN}. DeepCCA \cite{pmlr-v28-andrew13} shares the correlation maximizing objective with classic CCA, but learns a non-linear projection via deep neural networks. It has been shown to outperform linear CCA and its non-linear KCCA extension. In multi-modal DAEs \cite{ICML2011Ngiam_399} multi-modal autoencoders are trained with a shared hidden layer. More generally paired data has been used to train two branch DNNs to learn view-invariant embeddings for example via a learning to rank \cite{frome2013devise,Wang2017LearningTN} objective. 

In contrast to these Euclidean-based metrics, statistical dependency-based measures have like the Hilbert-schmidt independence criterion (HSIC) \cite{Gretton:2005:MSD:2101372.2101382} have been much less studied as objectives for multi-modal learning. One example is HSIC-CCA  \cite{chang13}, which learned a CCA type architecture but with HSIC rather than correlation objective.

However, the above supervised  algorithms -- particularly the deep learning ones -- require a large number of \emph{paired} samples to learn an effective cross-modal embedding.

\keypoint{Unsupervised multi-modal learning}
The desirability of learning from more widely available unpaired data has motivated some research into the harder problem of unsupervised cross-modal learning by introducing latent variables for cross-view pairing. An early approach was Matching CCA \cite{haghighi2008learning}. It alternates between learning a joint embedding space with CCA, and solving a bipartite matching problem to associate the unpaired data. Unlike the statistical dependency measures, CCA's correlation-based objective requires comparable embeddings to estimate a match. So Matching CCA can never bootstrap itself if initialised with completely random embeddings and no pairing information at all. Indeed it was only shown to work when used with a seed of paired samples for bootstrapping \cite{haghighi2008learning} -- i.e., in the semi-supervised setting. Probabilistic latent variable approaches have also been proposed to match across-views  \cite{iwata2013unsupervised}, however this was only demonstrated to work on toy problems. Both of these  are limited to linear projections.

To handle non-linearity in unsupervised multi-modal learning,  kernel based approaches were proposed including  Kernelized sorting (KS) \cite{djuric2012convex,Jeb04,quadrianto2009kernelized} and the least-squared object matching \cite{yamada2015cross,yamada2011cross}. In KS, unpaired data are matched by maximizing HSIC, and it has outperformed MCCA on NLP tasks \cite{jagarlamudi2010kernelized}. In LSOM, squared-loss mutual information (SMI) is used as a dependence measure, and it was experimentally shown to outperform the HSIC-based KS. However, both KS and LSOM are shallow methods, so may not perform well for image and text data where representation learning is beneficial. In this paper we leverage HSIC and SMI-based objectives for learning representations for matching in a deeper context.

\subsection{Applications}
\keypoint{Visual Description with Natural Language} Generating or matching natural language descriptions for images and videos has recently become a popular topic in cross-modal learning in the last five years \cite{ordonez2011im2text}. A common approach is to learn an image representation (e.g., CNN), a text representation (e.g., Bag of Words or LSTM \cite{Hochreiter:1997:LSM:1246443.1246450}) and then map these into a common latent space via two-branch deep networks \cite{klein2015captioningFV,Wang2017LearningTN,Yan_2015_CVPR}. Based on this latent space, images or videos and associated text descriptions can be matched: supporting annotation or retrieval applications.  Our proposed  ~\mabbr~~  solves  supervised image captioning  comparably well to the state of the art methods. But unlike prior approaches it can be generalized to the semi-supervised and unsupervised case for exploiting unpaired data.

\keypoint{Zero-shot learning}  Zero-shot learning (ZSL) aims to improve the scalability of visual classifier learning in terms of the required image annotation. Specifically, it assumes that a subset of image categories (`seen' categories) are provided with example images (i.e., annotations), and aims to generate classifiers for a disjoint set of image categories (`unseen' categories) that have no example images. This is achieved by further assuming that all image categories have an associated category \emph{embedding} vector, for example word-vectors \cite{mikolov2013distributedRepresentations} or attributes \cite{Lampert:2014:ACZ:2587733.2587824}. Given this assumption a cross-modal mapping can be learned from visual feature space to category embedding space using the data for seen classes. The cross-modal mapping can then be used as a classifier in order to recognise those unseen categories for which no annotated examples were provided \cite{frome2013devise,socher2013zslCrossModal}. 

Our ~\mabbr~ approach is related to ZSL methods in that it can be applied to learn cross-modal embeddings between images and category vectors, and hence it can also be used as a classifier for novel classes. However it has a few crucial benefits: (i) It can be learned in a semi-supervised way, which {encompasses the transductive \cite{fu2015multiviewZSL,DBLP:journals/corr/TsaiHS17} and semi-supervised \cite{DBLP:journals/corr/TsaiHS17} variants of ZSL}. (ii) More interestingly, it can also be learned in an entirely un-supervised way -- requiring \emph{no paired samples at all}; unlike all existing ZSL methods. {We term this specific problem setting ~\tname~ (UCL)}. 

A recent ZSL method ReViSE \cite{DBLP:journals/corr/TsaiHS17}  is related to ours in that it can also benefit from the semi-supervised learning setting {via a HSIC-based domain adaptation loss}. However ReViSE is engineered specifically for ZSL. In contrast our ~\mabbr~  is a general cross-modal learner, and can address the completely unsupervised setting unlike ReViSE.

\section{Matching Deep Autoencoders}
\label{sec:Matching Deep Autoencoders}
We introduce our cross-domain object matching methodology, \mname~ (Figure~\ref{fig:mdae-arch-2}) from the unsupervised learning perspective where no paired training data is assumed. From here semi-supervised and supervised variants are a straightforward special case. {For exposition simplicity we also assume an equal number of samples in each view, but this can be relaxed in practice.}

\subsection{Multi-View Autoencoders}
Consider two unpaired sets of samples, $\{\boldx_i\}_{i = 1}^n$ and $\{\boldy_i\}_{i = 1}^{{n}}$, where $\boldx \in \mathbbR^{d_x}$ and $\boldy \in \mathbbR^{d_y}$. For example, $\boldx$ is a \textit{feature vector} extracted from an image and $\boldy$ is a vector representation of a text. 
We assume a heterogeneous setup; the dimensionality of $\boldx$ and $\boldy$ are completely different.

Let us denote the autoencoders of $\boldx$ and $\boldy$ as
\begin{align*}
\boldf_{\boldx}(\boldg_{\boldx}(\boldx; \boldTheta_{\boldx}); \boldTheta_{\boldx}),~~~ \boldf_{\boldy}(\boldg_{\boldy}(\boldy;\boldTheta_{\boldy});\boldTheta_{\boldy}),
%\boldg_{\boldy}(\boldf_{\boldy}(\boldy)) = \boldTheta_{\boldy} \text{Relu}(\boldTheta_{\boldy}^\top \boldy),
\end{align*}
where $\boldg(\cdot)$ is an encoder and $\boldf(\cdot)$ is a decoder function, $\boldTheta_{\boldx}$, and $\boldTheta_{\boldy}$ are the autoencoder parameters. %$m$ is the dimensionality of the latent space, $\text{Relu}(\boldx) = \max(\boldzero, \boldx)$ is the element-wise rectified linear unit, and $\boldzero$ is the vector of all zeros. 

Our goal is to learn comparable representation embeddings $\bm g_x(\cdot)$ and $\bm g_y(\cdot)$ given no paired training data. This is a significantly harder problem than other multi-modal autoencoder approaches that rely on paired data \cite{chandar2016corrNNs,ICML2011Ngiam_399}.

%The final goal of this paper is to \emph{jointly} estimate model parameters $\boldTheta_{\boldx}$ and $\boldTheta_{\boldy}$ from \emph{unpaired data}. That is, we train a multi-vew DAEs from unpaired data.  Typically, for learning multi-modal DAEs, we need to have a large number of paired data. Thus, this problem is extremely difficult compared to standard multi-modal DAEs. In addition to the fully unsupervised setup, we challenge to solve a semi-supervised problems, where we have a small number of paired data and a large number of unpaired data. To the best of our knowledge, this is the first work to learn multi-modal DAEs from unpaired data (or less supervision).

\subsection{Learning from Unpaired Data}
To learn from unpaired data we introduce a permutation matrix to represent the unknown correspondence between data items in two views \cite{quadrianto2009kernelized,yamada2011cross,yamada2015cross}.  Let $\pi$ be an permutation function over $\{1, 2, \ldots, n\}$, and let $\boldPi$ be the corresponding permutation indicator matrix:
\begin{align*}
\boldPi \in \{0,1\}^{n \times n}, \boldPi \boldone_n = \boldone_n, \text{and}~ \boldPi^\top \boldone_n = \boldone_n,
\end{align*}
where $\boldone_n$ is the $n$-dimensional vector with all ones.

Then, we consider the following optimization problem:
\begin{align}
%&\widehat{\boldTheta}_{\boldx}, \widehat{\boldTheta}_{\boldy}, \widehat{\boldPi} \\
& \min_{\boldTheta_{\boldx}, \boldTheta_{\boldy}, \boldPi}  \sum_{i = 1}^n \|\boldx_i - \boldf_{\boldx}(\boldg_{\boldx}(\boldx_i))\|_2^2 + \|\boldy_i - \boldf_{\boldy}(\boldg_{\boldy}(\boldy_i)))\|_2^2\nonumber\\
&\phantom{\min_{\boldTheta_{\boldx}, \boldTheta_{\boldy}, \boldPi}}- \lambda D_\Pi(\{\boldg_{\boldx}(\boldx_i), \boldg_{\boldy}(\boldy_{\pi(i)})\}_{i= 1}^n)\label{eq:mainObj},
\end{align}
where we are simultaneously optimising the autoencoders ($\boldTheta_{\boldx}$ and $\boldTheta_{\boldy}$ ) as well as the cross-domain match ($\boldPi$) with tradeoff parameter $\lambda$. The key component here is the function $D_{\boldPi}(\cdot,\cdot)$ which is a non-negative statistical dependence measure between the $x$ and $y$ views. {$D_{\boldPi}(\cdot,\cdot)$ needs to be a measure which does not require comparable representations \emph{a priori} in order to enable learning to get started. }

\begin{figure}[t]
\begin{center}
\begin{minipage}[t]{0.99\linewidth}
\centering
  {\includegraphics[width=0.99\textwidth]{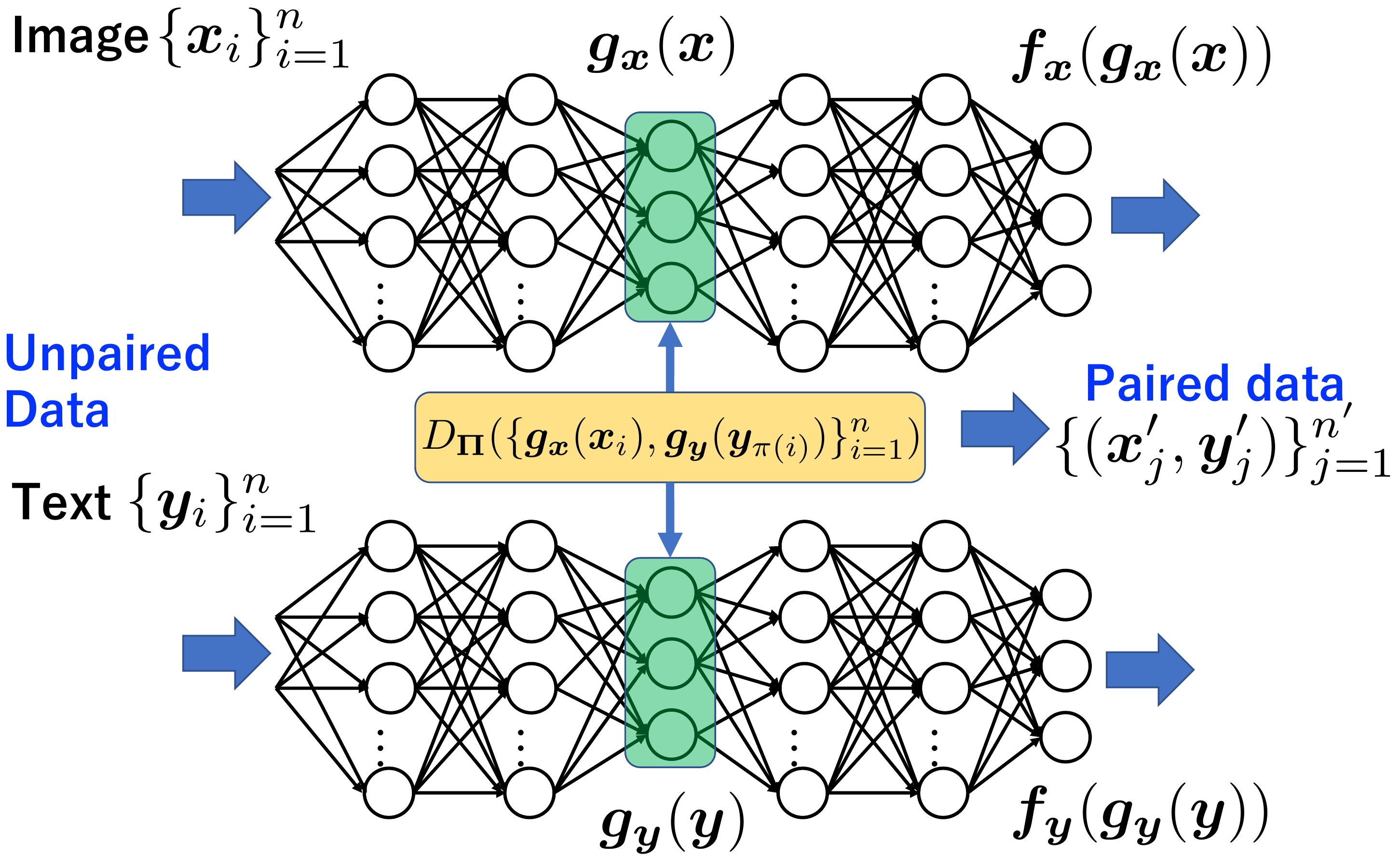}} \\ \vspace{-0.10cm}
\end{minipage}
\centering
\caption{Model Architecture of DMAE.}
\label{fig:mdae-arch-2}
\end{center}
\end{figure}

\subsection{Dependence Measures}
The statistical dependence measure is the crucial component in achieving our goal. In this paper, we explore two alternatives:  the unnormalized kernel target alignment (KTA) \cite{cristianini2002kernel} and the squared-loss mutual information (SMI) \cite{suzuki2010sufficient,yamada2015cross,yamada2011cross}. Note that SMI is an independence measure. However, since we want to make $\boldTheta_\boldx$ and $\boldTheta_\boldy$ generate similar representations, we use SMI as a dependence measure.

\keypoint{Unnormalized kernel target alignment (uKTA)}
In uKTA \cite{cristianini2002kernel}, we consider the following similarity function $D$ for paired data:
\begin{align*}
\text{uKTA}(\{(\boldx_i, \boldy_i)\}_{i = 1}^n) = \text{tr}\left(\boldK \boldL \right),
\end{align*}
where $\text{tr}(\cdot)$ is the trace operator,  $\boldK$ is the Gram matrix for $\boldx$ and $\boldL$ is the Gram matrix for $\boldy$. This similarity function takes large value if the Gram matrices $\boldK$ and $\boldL$ are similar, and a small value if they are not similar. Note that, in the original KTA, we have the normalization term. However, this makes the optimization hard, and thus we employ the unnormalized variant of KTA. Moreover, uKTA can be regarded as a non-centered variant of HSIC \cite{Gretton:2005:MSD:2101372.2101382}.

We use the Gaussian kernel:
\begin{align*}
\boldK_{ij} \!\!=\! \exp\left(-\frac{\|\boldx_i - \boldx_j\|_2^2}{2\sigma_x^2}\right), ~\boldL_{ij} \!\!=\! \exp\left(-\frac{\|\boldy_i - \boldy_j\|_2^2}{2\sigma_y^2}\right),
\end{align*}
where $\sigma_x > 0$ and $\sigma_y>0$ are the Gaussian width. 

Given un-aligned data which depends on a permutation matrix $\boldPi$ with respect to $\boldy$, we can write uKTA  as
\begin{align}
\text{uKTA}(\{(\boldx_i, \boldy_{\pi(i)})\}_{i = 1}^n) = \text{tr}\left(\boldK \boldPi^\top \boldL \boldPi \right).\label{eq:uKTAobj}
\end{align}
%\doublecheck{Note that, uKTA is not a dependence measure; it is not zero when two random variables are not independent. To handle this issue, we can use the Hilbert-Schmidt Independence Criterion (HSIC) \cite{Gretton:2005:MSD:2101372.2101382}.} 

\vspace{.1in}
\noindent {\bf Squared-Loss Mutual Information (SMI)}
The squared loss mutual information (SMI) 
%\doublecheck{}
between  two random variables 
is defined as \cite{suzuki2010sufficient}
\begin{align*}
\text{SMI} = \iint \left(\frac{p(\boldx,\boldy)}{p(\boldx)p(\boldy)} - 1 \right)^2 p(\boldx)p(\boldy) \text{d}\boldx \text{d} \boldy,
\end{align*}
which is the Pearson divergence \cite{doi:10.1080/14786440009463897} from
$p(\boldx,\boldy)$ to $p(\boldx)p(\boldy)$. The SMI is an $f$-divergence \cite{10.2307/2984279} that is it is a non-negative measure and is zero only if the random variables are independent.   
%$where$ 
%it takes zero value if the random variables are independent, and takes non-negative value if it is not independent. %\thnote{Big $X$ vs small $x$?} 
%\mynote{Big $X$ is random variable and $\boldx$ is an observed vector. However, since we are not using $X$, I simply get rid of it.}

To estimate the SMI, a direct density ratio estimation approach is useful. The key idea is to approximate the true density ratio by the model:
\begin{align*}
r(\boldx,\boldy;\boldalpha) = \sum_{\ell = 1}^n \alpha_\ell K(\boldx_\ell, \boldx)L(\boldy_\ell, \boldy),
\end{align*}
where $\boldalpha = [\alpha_1, \ldots, \alpha_n]^\top \in \mathbbR^n$ is the model parameter. 

Then, the model parameter is given by minimizing the error between true density-ratio and its model:
\begin{align*}
J(\boldalpha) = \iint \left(\frac{p(\boldx,\boldy)}{p(\boldx)p(\boldy)} - r(\boldx,\boldy;\boldalpha)\right)^2 p(\boldx)p(\boldy) \textnormal{d}\boldx \textnormal{d}\boldy.
\end{align*}
%where $\boldalpha = [\alpha_1, \ldots, \alpha_n]^\top \in \mathbbR^n$.

By approximating the loss function by samples, we have the following optimization problem \cite{suzuki2010sufficient}:
\begin{align*}
\min_{\boldalpha} &\hspace{0.3cm} \frac{1}{2}\boldalpha^\top \widehat{\boldH} \boldalpha - \boldalpha^\top \widehat{\boldh} + \frac{\lambda}{2} \|\boldalpha\|_2^2,
\end{align*}
where
\begin{align*}
%\widehat{\boldalpha} &= \left(\widehat{\boldH} + \lambda \boldI_n\right)^{-1} \boldh, \\
\widehat{\boldH} &= \frac{1}{n^2} (\boldK\boldK^\top)\circ(\boldL\boldL^\top),~~~ \widehat{\boldh} = \frac{1}{n}(\boldK \circ \boldL)\boldone_n,
\end{align*}
$\lambda \geq 0$ is a regularization parameter and $\circ$ is the elementwise product.% and $\boldone_n$ is the $n$-dimensional vector whose element are all ones. 

The optimal solution of the above optimization problem can be analytically obtained
\begin{align*}
\widehat{\boldalpha} &= \left(\widehat{\boldH} + \lambda \boldI_n\right)^{-1} \widehat{\boldh},
\end{align*}
where $\boldI_n$ is the $n\times n$ dimensional identity matrix. 

Then, the estimator of SMI can be given as \cite{yamada2015cross,yamada2011cross}:
\begin{align*}
\widehat{\text{SMI}}(\{(\boldx_i, \boldy_i)\}_{i = 1}^n) = \frac{1}{2n} \text{tr}\left( \diag{\widehat{\boldalpha}}\boldK\boldL\right) - \frac{1}{2},
\end{align*}
where $\diag{\boldalpha} \in \mathbbR^{n \times n}$ is the diagonal matrix whose diagonal elements are $\boldalpha$.  We can see that uKTA is a special case of SMI. Specifically, if we set $\widehat{\boldalpha} = \boldone_n$, SMI boils down to uKTA. 
 
To extend SMI-based dependency to the permuted case, we can  replace  $\boldL \rightarrow \boldPi^\top \boldL \boldPi$:
\begin{align*}
\widehat{\text{SMI}}(\{(\boldx_i, \boldy_{\pi(i)})\}_{i = 1}^n) \!=\! \frac{1}{2n} \text{tr}\left( \diag{\widehat{\boldalpha}_{\boldPi}}\boldK \boldPi^\top \boldL \boldPi \right)\!-\! \frac{1}{2}.
\end{align*}

\subsection{Optimization}
For initializing $\boldTheta_{\boldx}$ and $\boldTheta_{\boldy}$, we first independently estimate $\boldTheta_{\boldx}$ and $\boldTheta_{\boldy}$ by using autoencoders.  Then we employ an alternative optimization for learning $\boldTheta_{\boldx}$ and $\boldTheta_{\boldy}$ and $\boldPi$ together. We optimize $\boldTheta_{\boldx}$ and $\boldTheta_{\boldy}$ with fixed $\boldPi$, and then optimize $\boldPi$ with fixed $\boldTheta_{\boldx}$ and $\boldTheta_{\boldy}$. This alternation is continued until convergence.

\keypoint{Optimization for $\boldTheta_{\boldx}$ and $\boldTheta_{\boldy}$}
With fixed permutation matrix ${\boldPi}$ (or ${\pi}$), the objective function is written as
\begin{align*}
 \min_{\boldTheta_{\boldx}, \boldTheta_{\boldy}} &\sum_{i = 1}^n \|\boldx_i - \boldf_{\boldx}(\boldg_{\boldx}(\boldx_i))\|_2^2 + \|\boldy_i - \boldf_{\boldy}(\boldg_{\boldy}(\boldy_i)))\|_2^2\\
& -\lambda D_\boldPi(\{\boldg_{\boldx}(\boldx_i), \boldg_{\boldy}(\boldy_{{\pi}(i)})\}_{i= 1}^n).
\end{align*}

This problem is an autoencoder optimization with regularizer $D_\boldPi(\cdot,\cdot)$, and can be solved with backpropagation.

\vspace{.1in}
\noindent {\bf Optimization for $\boldPi$ (KTA)} For optimizing $\boldPi$, we employ a kernelized sorting \cite{djuric2012convex,quadrianto2009kernelized} strategy.

The empirical estimate of uKTA using $\{\boldg_{\boldx}(\boldx_i), \boldg_{\boldy}(\boldy_{\pi(i)})\}_{i = 1}^n$ is given as
\begin{align}
\label{eq:ks}
\text{uKTA} = \text{tr}\left(\boldK_{\boldTheta_{\boldx}}\boldPi^\top\boldL_{\boldTheta_{\boldy}}\boldPi\right),
\end{align}
where
\begin{align*}
[\boldK_{\boldTheta_{\boldx}}]_{ij} &= \exp\left(-\frac{\|\boldg_{\boldx}(\boldx_i) - \boldg_{\boldx}(\boldx_j)\|_2^2}{2\sigma_x^2}\right), \\
[\boldL_{\boldTheta_{\boldy}}]_{ij} &= \exp\left(-\frac{\|\boldg_{\boldy}(\boldy_i) - \boldg_{\boldy}(\boldy_j)\|_2^2}{2\sigma_y^2}\right).
\end{align*}
%For the distance covariance, we simply need to change the kernel to the distance. 

\noindent The optimization problem can then be written as:
\begin{align*}
\max_{\boldPi \in \{0,1\}^{n\times n}} %& \hspace{0.3cm} \|\boldY - \boldTheta_{\boldy}\text{Relu}(\boldTheta_{\boldy}\boldY)\boldPi\|_{F}^2 \\
&\hspace{0.3cm} \text{tr}\left(\boldK_{\boldTheta_{\boldx}}\boldPi\boldL_{\boldTheta_{\boldy}}\boldPi^\top \right) \\
\text{s.t.} & \hspace{0.3cm} \boldPi \boldone_n = \boldone_n, \boldPi^\top \boldone_n = \boldone_n.
\end{align*}

This problem is known as quadratic assignment programming and NP-hard. Thus, we relax the assignment matrix $\boldPi$ to take real values as
\begin{align*}
\boldPi \in [0, 1]^{n \times n}, \boldPi \boldone_n = \boldone_n, \boldPi^\top \boldone_n = \boldone_n.
\end{align*}

%\doublecheck{With the relaxation, Eq.~\eqref{eq:ks} becomes convex. However, since we  maximize $\boldPi$ over convex function Eq.~\eqref{eq:ks}, we can get poor locally optimal solution due to bad initialization. To solve the problem, we employ the convex optimization proposed by \cite{djuric2012convex}.} \thnote{Unclear}

\begin{lemm} \cite{djuric2012convex}
\label{lemm1}
\begin{align*}
\argmax_{\boldPi} \textnormal{tr}\left(\boldK\boldPi^\top \boldL \boldPi\right) \!=\! \argmin_{\boldPi} \|\boldK\boldPi^\top \!-\! (\boldL \boldPi)^\top \|_{F}^2.
\end{align*}
where $\|\boldA\|_{F}$ is Frobenius norm and $\boldPi \in \{0,1\}^{n\times n}$.
\end{lemm}

Using the continuous relaxation of $\boldPi$ and Lemma~\ref{lemm1}, we can write the optimization function as %\cite{djuric2012convex}
\begin{align*}
\min_{\boldPi \in [0,1]^{n\times n}} %& \hspace{0.3cm} \|\boldY - \boldTheta_{\boldy}\text{Relu}(\boldTheta_{\boldy}\boldY)\boldPi\|_{F}^2 \\
&\hspace{0.3cm} \|\boldK_{\boldTheta_{\boldx}}\boldPi^\top - (\boldL_{\boldTheta_{\boldy}}\boldPi)^\top \|_F^2 \\
\text{s.t.} & \hspace{0.3cm} \boldPi \boldone_n = \boldone_n, \boldPi^\top \boldone_n = \boldone_n.
\end{align*}
This problem is convex with respect to $\boldPi$, and thus, we can obtain a globally optimal solution for this sub-problem. 

To efficiently estimate the permutation matrix, we employ the regularization based optimization technique \cite{djuric2012convex}:
\begin{align*}
\min_{\boldPi} &\hspace{0.3cm} \|\boldK_{\boldTheta_{\boldx}}\boldPi^\top - (\boldL_{\boldTheta_{\boldy}}\boldPi)^\top \|_F^2 \\
&\hspace{0.3cm} +\lambda_{\boldPi} \sum_{k = 1}^n \left((\sum_{\ell = 1}^n \boldPi_{k\ell} -1)^2 + (\sum_{\ell=1}^n\boldPi_{\ell k} -1)^2\right)\\
\text{s.t.} & \hspace{0.3cm} \boldPi_{k\ell} \geq 0, \text{for}~k,\ell \in \{1,2,\ldots, n\},
\end{align*}
where $\lambda_{\boldPi} \geq 0$ is the regularization parameter. The $\boldPi$ can be optimized by using gradient descent. 

\keypoint{Optimization for $\boldPi$ (SMI)} For optimizing $\boldPi$, we employ a regularized variant of  LSOM \cite{yamada2015cross,yamada2011cross}.

The empirical estimate of SMI using $\{\boldg_{\boldx}(\boldx_i), \boldg_{\boldy}(\boldy_{\pi(i)})\}_{i = 1}^n$ is given as
\begin{align}
\label{eq:smi}
\text{SMI} = \frac{1}{2n}\text{tr}\left(\diag{\widehat{\boldalpha}_{\boldTheta}}\boldK_{\boldTheta_{\boldx}}\boldPi^\top\boldL_{\boldTheta_{\boldy}}\boldPi\right) - \frac{1}{2}.
\end{align}

The optimization problem can be given as
\begin{align*}
\max_{\boldPi} &\hspace{0.3cm} \text{tr}\left(\diag{\widehat{\boldalpha}_{\boldTheta}}\boldK_{\boldTheta_{\boldx}}\boldPi^\top\boldL_{\boldTheta_{\boldy}}\boldPi\right) \\
&\hspace{0.3cm} +\lambda_{\boldPi} \sum_{k = 1}^n \left((\sum_{\ell = 1}^n \boldPi_{k\ell} -1)^2 + (\sum_{\ell=1}^n\boldPi_{\ell k} -1)^2\right)\\
\text{s.t.} & \hspace{0.3cm} \boldPi_{k\ell} \geq 0, \text{for}~k,\ell \in \{1,2,\ldots, n\}
%\text{s.t.} & \hspace{0.3cm} \boldPi \boldone_n = \boldone_n, \boldPi^\top \boldone_n = \boldone_n.
\end{align*}
This optimization problem can be solved by using gradient ascent. 

%To solve this, we employ a regularization based LSOM algorithm \cite{yamada2011cross,yamada2015cross}. %, in which we alternatively optimize $\widehat{\boldalpha}_{\boldTheta}$ and $\boldPi$. Note that, since $\boldPi$ is included in $\widehat{\boldalpha}_{\boldTheta}$, we compute $\widehat{\boldalpha}_{\boldTheta}$ with previous $\boldPi$, and then optimize $\boldPi$ with fixing $\widehat{\boldalpha}_{\boldTheta}$. This is a heuristic approach, however, it in practice performs well.
%\mynote{Tanmoy, please confirm the training w.r.t. SMI measure. }

\subsection{Learning from Paired and Unpaired Data} \label{subsection:Semi-supervised data}
In the previous section we introduced our method assuming no paired data was available (unsupervised) case. In this section we explain our method in the case of some paired data (semi-supervised) case. 

Denote the paired data as $\{(\boldx'_j, \boldy'_j)\}_{j = 1}^{n'}$ and unpaired data as $\{\boldx_i\}_{i = 1}^n$ and $\{\boldy_i\}_{i = 1}^n$ ($n' < n$). Then, the  semi-supervised variant of ~\mabbr~~is as follows:

\keypoint{Optimization for $\boldTheta_{\boldx}$ and $\boldTheta_{\boldy}$}
With fixed permutation matrix ${\boldPi}$ (or ${\pi}$), the objective function is written as
\begin{align*}
%&\widehat{\boldTheta}_{\boldx}, \widehat{\boldTheta}_{\boldy}, \widehat{\boldPi} \\
 \min_{\boldTheta_{\boldx}, \boldTheta_{\boldy}} &\sum_{i = 1}^n \|\boldx_i - \boldf_{\boldx}(\boldg_{\boldx}(\boldx_i))\|_2^2 + \|\boldy_i - \boldf_{\boldy}(\boldg_{\boldy}(\boldy_i)))\|_2^2\\
& + \sum_{j = 1}^{n'} \|\boldx'_j - \boldf_{\boldx}(\boldg_{\boldx}(\boldx'_i))\|_2^2 + \|\boldy'_j - \boldf_{\boldy}(\boldg_{\boldy}(\boldy'_j)))\|_2^2\\
& -\lambda D_\boldPi(\{\boldg_{\boldx}(\boldx_i), \boldg_{\boldy}(\boldy_{{\pi}(i)})\}_{i= 1}^n)\\
& -\lambda D(\{\boldg_{\boldx}(\boldx'_i), \boldg_{\boldy}(\boldy'_{j})\}_{j= 1}^{n'}).
\end{align*}
%Here, $\boldI_{n'} \in \mathbbR^{n' \times n'}$ is the $n' \times n'$ dimensional identity matrix for paired data.% and ${\boldPi} \in \mathbbR^{n-n' \times n-n'}$ is the permutation matrix for unpaired data. 
 This is optimized with backpropagation as before.

\keypoint{Optimization for $\boldPi_{n}$} With given $\boldTheta_{\boldx}$ and $\boldTheta_{\boldy}$, we optimize $\boldPi$. 

\begin{align*}
\max_{\boldPi} & D_{\boldPi}(\{\boldg_{\boldx}(\boldx_i), \boldg_{\boldy}(\boldy_{{\pi}(i)})\}_{i= 1}^n)\\
&\hspace{0.3cm} +\lambda_{\boldPi} \sum_{k = 1}^n \left((\sum_{\ell = 1}^n \boldPi_{k\ell} -1)^2 + (\sum_{\ell=1}^n\boldPi_{\ell k} -1)^2\right)\\
\text{s.t.} & \hspace{0.3cm} \boldPi_{k\ell} \geq 0, \text{for}~k,\ell \in \{1,2,\ldots, n\}.
%\text{s.t.} & \hspace{0.3cm} \boldPi \boldone_n = \boldone_n, \boldPi^\top \boldone_n = \boldone_n. 
%& \hspace{0.3cm} {\boldPi} = \left[
%\begin{array}{cc}
%\boldI_{n'} & \boldO \\
%\boldO &  {\boldPi}_{n-n'}
%\end{array}
%\right].
\end{align*}

\keypoint{Fully Supervised Case} 
The fully supervised case is a trivial extension of the above. In this case $n=0$, $\boldPi$ is given, and we only need to optimize $\boldTheta_x$ and $\boldTheta_y$ for matching criterion $D_\boldPi(\cdot,\cdot)$.

\keypoint{Many-one-Pairing}
We introduced the methodology for 1-1 pairing with a square $\bm\Pi$ matrix. We can relax this assumption to obtain many-one pairing by considering rectangular $\bm\Pi $ and removing the column-sum constraint.

\section{Experiments}
\label{sec:Experiments}

We evaluate our contributions with two sets of experiments including image-caption matching (Section~\ref{sec:caption}) and classifier learning (Section~\ref{subsection:Unsupervised Classifier learning}).

\keypoint{Alternatives: Semi-supervised}
We evaluate our proposed {\bf DMAE-uKTA} and {\bf DMAE-SMI} methods against the following alternatives for unpaired data learning:\\
\textbf{MCCA:}\quad Matching CCA \cite{haghighi2008learning} for learning from paired and unpaired data across multiple views.\\
\textbf{Deep-MCCA:}\quad The original Matching CCA \cite{haghighi2008learning} is a shallow method. We extend it to a multi-layer deep architecture.\\
{\bf Deep-uKTA:}\quad  DMAE-uKTA without reconstruction loss. \\
{\bf Deep-SMI:}\quad Our DMAE-SMI without reconstruction loss. 

\keypoint{Alternatives: Supervised}
For supervised learning, we evaluate the following state of the art alternatives:\\
{\bf DeepCCA:} CCA with deep architecture \cite{pmlr-v28-andrew13}. \\
{\bf HSIC-DeepCCA:} A baseline we created. Extending HSIC-CCA \cite{chang13} to include a deep architecture, or updating DeepCCA to use HSIC rather than correlation-based loss.\\
{\bf Two-way Nets:} Two way nets use pre-trained VGG networks followed by fully connected layers (FC) and ReLU nonlinearities \cite{wang2016imageTextEmbed,Wang2017LearningTN}. Captioning only.   \\
{\bf ReViSE:} uses  autoencoders for each modality, minimizing the reconstruction loss for each modality and also the maximum mean discrepancy between them \cite{DBLP:journals/corr/TsaiHS17} . ZSL only.

%In Table~\ref{table:Differences between methods} we highlight the differences between various supervised and semi-supervised versions. 

\keypoint{Settings} We implement DMAE with Theano. The number of encoding and decoding layers were set to 3 (See Figure~\ref{fig:mdae-arch-2} for the model architecture). The regularization parameter were  set to $\lambda = 0.7$, $\lambda_{\boldPi} = 1.0$, and the kernel parameters $\sigma_x^2 = 2.5$ and $\sigma_y^2 = 0.5$ for all experiments. The experiments were run on NVIDIA P100 GPU processors. 

%\begin{figure}[t]
%\begin{center}
%\centering
%  {\includegraphics[width=0.99\columnwidth]{Average_precision.pdf}} 
%\caption{Average precision of the \doublecheck{DMAE-SMI} method over 10 classes from Animals with Attributes.}
%\label{fig:avg_precision}
%\end{center}
%\end{figure}

\subsection{Image-Sentence/Sentence-Image Retrieval}\label{sec:caption}
\keypoint{Benchmark Details} We evaluate Image$\to$Sentence and Sentence$\to$Image retrieval using the widely studied Flickr30K \cite{TACL229} and  MS-COCO \cite{Lin2014} datasets. Flickr30K consists of 31,783 images accompanied by descriptions. The larger MS-COCO dataset \cite{Lin2014} consists of 123,000 images, along accompanied by descriptions. Each dataset has 1000 testing images. Flickr30K has 5000 test sentences and COCO has 1000. To compare the methods, we use the evaluation metrics proposed in \cite{7534740}: Image-text and text-image matching performance quantified by Recall@$K=\{1,5\}$. We encode each image with $4096d$ VGG-19 deep feature  \cite{simonyan2014very} and a 300$d$ word-vector \cite{mikolov2013distributedRepresentations} average for each sentence.
%\todo{@TK: Need detail of sentence embedding and image embedding. Type and dimension (VGG19 and 300D WV)}.

\keypoint{Supervised Learning}  In the first experiment we evaluate our methods against prior state of the art in Image-Sentence matching under the standard supervised learning setting. From the results in Table~\ref{tab:captionFSL} we make the following observations: (i) Our SMI provides a better objective for our method than uKTA, this is expected since as we saw uKTA is a special case of SMI. (ii) Our autoencoding approach is helpful for cross-view representation learning as DMAE-SMI and DMAE-uKTA outperform Deep-SMI and Deep-uKTA respectively. (iii) Overall our approach performs comparably or slightly better than alternatives in this fully supervised image-caption matching setting. This is despite that the most competitive alternative \cite{wang2016imageTextEmbed,Wang2017LearningTN} uses a more advanced Fisher-vector representation of text domain.

\begin{table}[t]
\centering
\caption{Fully supervised image-sentence matching results on Flickr30K and MS-COCO. $^1$ Our improved fusion of HSIC-CCA \cite{pmlr-v28-andrew13} and Deep-CCA \cite{chang13} $^2$ Our implementation of ReViSE$^b$ variant (reconstruction loss and MMD). Top block: Prior methods. Middle block: Ablations of our method. Bottom block: Our methods.\label{tab:captionFSL}}
%\resizebox{0.86\textwidth}{!}{
\begin{tabular}{lcccc}
\hline
\multicolumn{5}{c}{Flickr30K} \tabularnewline
\hline 
 %& \multicolumn{2}{c}{Image-to-Text} 
 & \multicolumn{2}{c}{Image-to-Text}  & \multicolumn{2}{c}{Text-to-Image}\tabularnewline
\hline 
Approach & R@1 & R@5  & R@1 & R@5 \tabularnewline
\hline 
DeepCCA \cite{pmlr-v28-andrew13} & 29.3 & 57.4 &   28.2 & 54.7 \tabularnewline
HSIC-DeepCCA$^1$ & 34.2 & 64.4 & 28.7 & 56.8 \tabularnewline
Two-way nets \cite{Wang2017LearningTN} & 49.8 & 67.5 & 36.0 & 55.6 \tabularnewline
ReViSE \textsuperscript{b} \cite{DBLP:journals/corr/TsaiHS17} $^2$ & 34.7 & 63.2 & 29.2 & 58.0 \tabularnewline
MCCA  \cite{haghighi2008learning} & 17.3 & 35.2  & 18.3 & 30.3  \tabularnewline
Deep-MCCA  & 27.8 & 37.7 & 24.0 & 28.4  \tabularnewline
\hline
Deep-uKTA  & 28.7 & 52.6 & 26.0 & 50.1 \tabularnewline
Deep-SMI  & 34.2 & 65.3 & 29.6 & 57.8 \tabularnewline
\hline 
DMAE-uKTA  & 47.4 & 63.7 & 32.7 & 51.4   \tabularnewline
DMAE-SMI  & \textbf{50.1} & \textbf{70.4} & \textbf{37.2} & \textbf{59.8} \tabularnewline
\hline 
\hline 
\multicolumn{5}{c}{MS-COCO} \tabularnewline
\hline 
DeepCCA \cite{pmlr-v28-andrew13} & 40.2 & 68.7 &   27.8 & 58.9 \tabularnewline
HSIC-DeepCCA$^1$  & 44.2 & 72.1 & 31.2 & 64.6 \tabularnewline
Two-way nets \cite{Wang2017LearningTN} & \textbf{55.8} & 75.2 & \textbf{39.7} & 63.3 \tabularnewline
ReViSE \cite{DBLP:journals/corr/TsaiHS17} & 51.8 & 76.3 & 38.7 & 64.2 \tabularnewline
MCCA  \cite{haghighi2008learning}  & 21.4 & 30.2 & 15.7 & 26.3 \tabularnewline
Deep-MCCA  & 25.6 & 30.2 & 19.8 & 27.0   \tabularnewline
\hline
Deep-uKTA  & 40.4 & 67.8 & 27.3 & 58.6 \tabularnewline
Deep-SMI  & 51.4 & 74.6 & 36.8 & 62.6 \tabularnewline
\hline 
DMAE-uKTA    & 49.8 & 68.2 & 37.2 & 65.2  \tabularnewline
DMAE-SMI  & 54.2 & \textbf{78.6} & 38.4 & \textbf{68.4}   \tabularnewline
\hline
\end{tabular}
\end{table}

\begin{table*}
\resizebox{1.0\textwidth}{!}{
\begin{tabular}{|c|cc|cc|cc|cc|cc|cc|cc|cc|}
\hline
\multicolumn{16}{c}{Flickr30K}\tabularnewline
\hline 
 & \multicolumn{8}{c|}{Supervised} & \multicolumn{8}{c|}{Un/Semi-Supervised}\tabularnewline
\hline 
 & \multicolumn{2}{c|}{MCCA} & \multicolumn{2}{c|}{Deep-MCCA} & \multicolumn{2}{c|}{DMAE-uKTA} & \multicolumn{2}{c|}{DMAE-SMI} & \multicolumn{2}{c|}{MCCA} & \multicolumn{2}{c|}{Deep-MCCA} & \multicolumn{2}{c|}{DMAE-uKTA} & \multicolumn{2}{c|}{DMAE-SMI}\tabularnewline
\hline 
Labels & I$\to$T & T $\to$ I & I $\to$ T & T $\to$ I & I $\to$ T & T $\to$ I & I $\to$T & T $\to$ I & I $\to$ T & T$\to$I & I$\to$T & T$\to$I & I$\to$T & T$\to$I & I$\to$T & T$\to$I\tabularnewline
\hline 
0\% & - & - & - & - & -& -& - & - & 1.2 & 0.8 & 1.8 & 1.2 & 4.2 & 2.3 & {\bf 4.9} & {\bf 3.0} \tabularnewline
20\% &  5.2 & 2.5 & 7.8 & 5.3 & 18.7 & 16.0 & 21.2 & 22.8 & 7.3 & 2.8 & 11.7 & 5.8 & 21.3 & 17.8 & {\bf 25.2} & {\bf 23.3}\tabularnewline
40\% & 8.9 & 12.0 & 21.8 & 16.2 & 30.8 & 22.8 & 32.7 & 29.6 & 10.2 & 12.8 & 23.4 & 17.0 & 32.8 & 24.3 & {\bf 35.3} & {\bf 31.8}\tabularnewline
100\% & 17.3 & 18.3 & 27.8 & 24.0 & 47.4 & 32.7 & {\bf 50.1}  & {\bf 37.2} &17.3 & 18.3 & 27.8 & 24.0 & 47.4 & 32.7 & {\bf 50.1}  & {\bf 37.2}\tabularnewline
\hline 
\end{tabular}}\\
\resizebox{1.0\textwidth}{!}{
\begin{tabular}{|c|cc|cc|cc|cc|cc|cc|cc|cc|}
\hline
\multicolumn{16}{c}{MS-COCO}\tabularnewline
\hline 
 & \multicolumn{8}{c|}{Supervised} & \multicolumn{8}{c|}{Un/Semi-Supervised}\tabularnewline
\hline 
 & \multicolumn{2}{c|}{MCCA} & \multicolumn{2}{c|}{Deep-MCCA} & \multicolumn{2}{c|}{DMAE-uKTA} & \multicolumn{2}{c|}{DMAE-SMI} & \multicolumn{2}{c|}{MCCA} & \multicolumn{2}{c|}{Deep-MCCA} & \multicolumn{2}{c|}{DMAE-uKTA} & \multicolumn{2}{c|}{DMAE-SMI}\tabularnewline
\hline 
Labels & I$\to$T & T $\to$ I & I $\to$ T & T $\to$ I & I $\to$ T & T $\to$ I & I $\to$T & T $\to$ I & I $\to$ T & T$\to$I & I$\to$T & T$\to$I & I$\to$T & T$\to$I & I$\to$T & T$\to$I\tabularnewline
\hline 
0\% & - & - & - & - & - & - & - & - & 0.7 & 0.3 & 1.2 & 0.8 & 1.8 & 1.0 & {\bf 2.2} & {\bf 1.6} \tabularnewline
20\% &  4.0 & 2.3 & 5.2 & 3.0 & 11.8 & 8.0 & 14.4 & 11.6 & 4.6 & 3.3 & 6.0 & 3.8 & 12.4 & 10.8 & {\bf 15.6} & {\bf 12.8}\tabularnewline
40\% & 10.1 & 12.0  & 15.8 & 16.2 & 21.8 & 18.8 & 26.8 & 24.8 & 15.3 & 10.2 & 18.4 & 11.3 & 22.4 & 19.7 & {\bf 28.4} & {\bf 25.6} \tabularnewline
100\% & 21.4 & 15.7 & 25.6 & 19.8 & 49.8 & 37.2 & {\bf 54.2} & {\bf 38.4} & 21.4 & 15.7 & 25.6 & 19.8 & 49.8 & 37.2 & {\bf 54.2} & {\bf 38.4}\tabularnewline
\hline
\end{tabular}}
\caption{Semi-supervised and unsupervised image-sentence retrieval results on Flickr30K and MS-COCO. R@1 metric.  
\label{tab:captionSSLnew}}
\end{table*}

\keypoint{Semi-supervised and Unsupervised Learning} In the second experiment we investigate whether it is possible to learn captioning from partially paired or unpaired data. For the results in Table~\ref{tab:captionSSLnew} the left (Supervised) block uses only the specified \% of labeled data, and the right (Un/Semi-supervised)  block uses both labeled and the available unlabeled data. 
We make the following observations: (i) As expected the task is clearly significantly harder as the percentage of labeled data decreases towards zero (rapid drop for methods under the supervised condition, left block). (ii) Our DMAE, and particularly DMAE-SMI perform effective semi-supervised learning (right), exploiting the unlabelled data to outperform the supervised only case (left), and performance decreases less rapidly as the \% of labeled data decreases. (iii) 
DMAE, and particularly our DMAE-SMI uses the unlabeled data much more effectively than prior MCCA \cite{haghighi2008learning} and our Deep-MCCA extension, outperforming them on semi-supervised learning at each evaluation point. (iv) DMAE-SMI performs significantly above chance (0.1\%) in completely \emph{unsupervised} image-caption matching.

\subsection{Unsupervised Classifier Learning}\label{subsection:Unsupervised Classifier learning}
We consider training a classifier given a stack of images and stack of category embeddings (we use word vectors) that describe the categories covered by images in the stack.

This is the `\tname~'  problem when there are \emph{no annotated images}, so no pairings given.
If the category labels of \emph{some} images are unknown, and all categories have at least one annotated image, this a semi-supervised learning problem. In the case where the category labels of \emph{some} images are unknown and some categories have no annotated images, this is a zero-shot learning problem. 
If category labels of all images are known (all pairings given), this is the standard supervised learning problem. Our framework can apply to all of these settings, but {as fully supervised and zero-shot learning are well studied, we focus on the unsupervised and semi-supervised variants}.

\keypoint{Benchmark Details} We evaluate our approach on AwA \cite{lampert2009zeroshot_dat} and CIFAR-10 \cite{krizhevsky2009learning} datasets.  As category embeddings, we  use $300d$ word-vectors \cite{mikolov2013distributedRepresentations}. For image features we use $4096d$ VGG-19 \cite{simonyan2014very} features for AwA, and  the $12800d$ feature  of \cite{ICML2011Coates_485} for CIFAR-10. Thus for AwA, image  data is a $4096 \times n$ stack of $n$ images, and  category domain data  is a $300\times m$ stack of $m=50$ word vectors. Unsupervised DMAE learns a joint embedding and the association matrix $\boldPi \in \{0,1\}^{n \times m}$  that pairs images with categories. The learned  $\boldPi$ should ideally match the 1-hot label matrix that would normally be given as a target in supervised learning. 

\keypoint{Settings} We consider three settings: 1. Unsupervised learning: No  paired data are given. 2. Semi-supervised learning: A subset of paired data are given, and the remaining unpaired data are to be exploited. 3. Transductive semi-supervised: As for semi-supervised, but the model can also see the (unpaired) testing data during training.

\keypoint{Metrics} To fully diagnose the performance, we evaluate a few metrics: (i) Matching accuracy. The accuracy of predicted $\widehat{\boldPi}$ on the training split compared to the ground-truth $\boldPi$ as quantified by average precision and average recall. (ii) Classifier accuracy. Using $\widehat{\boldPi}$ as labels to train a SVM classifier, we evaluate the accuracy of image recognition on the testing split using the trained SVM.

\keypoint{Results: Unsupervised Matching}  In the unsupervised classifier learning setting, it is a non-trivial achievement to correctly estimate associations between images and categories better than chance since we have no annotated pairings, and the domains are not a priori comparable. To quantify the accuracy of pairing, we compare estimated $\widehat{\boldPi}$ and true ${\boldPi}$ and compute compute the  precision and recall by class. After learning DMAE-SMI on AwA we obtain an impressive precision of \textbf{$0.90$} and recall of \textbf{$0.84$} averaged over all 50 classes given \emph{no prior pairings} to start with.

\begin{figure}[t]
 \begin{center}
 \begin{minipage}[t]{0.49\linewidth}
 \centering
   {\includegraphics[width=0.99\textwidth]{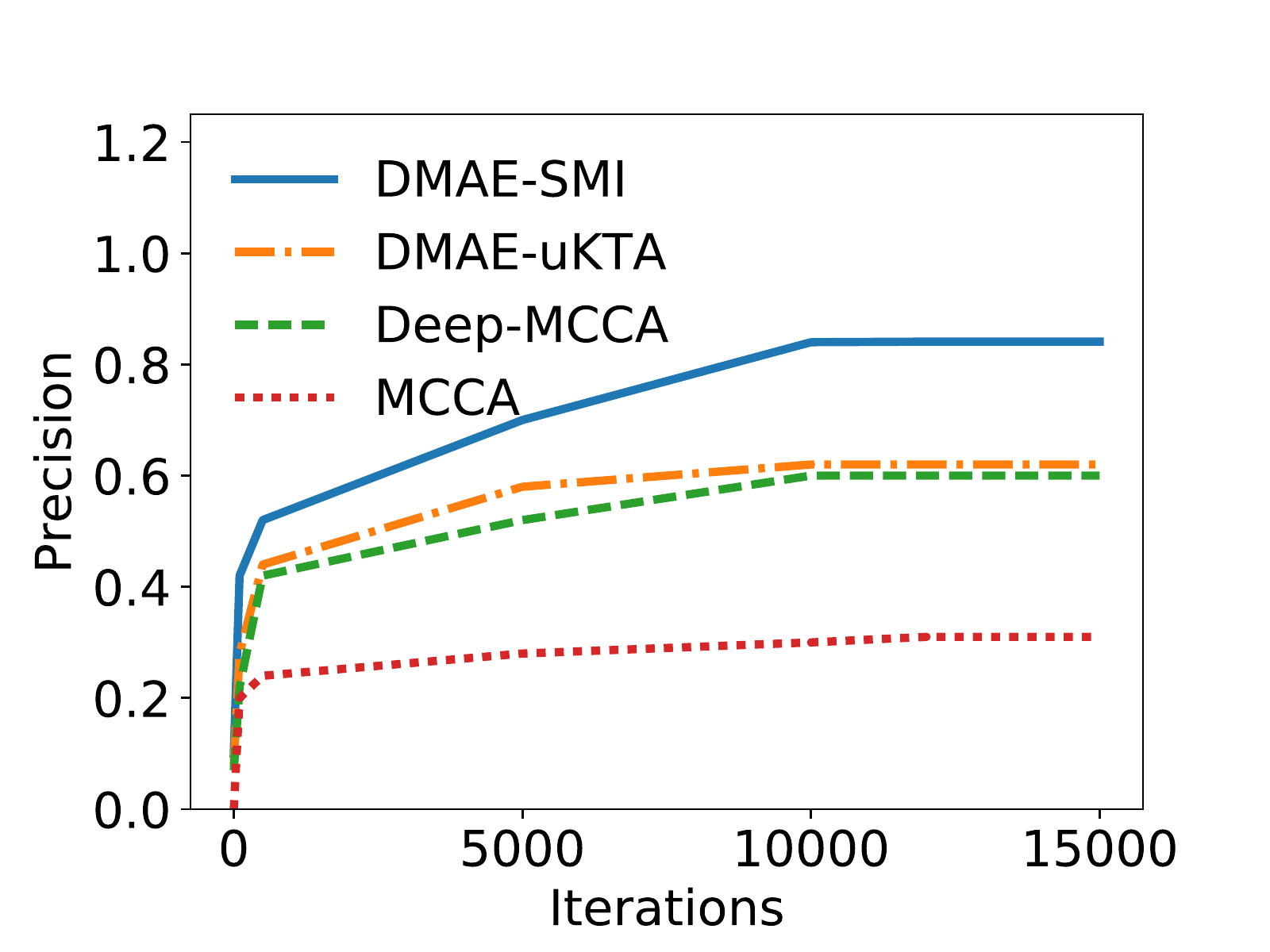}} \\% \vspace{-0.10cm}
   (a) Precision.
 \end{minipage}
 \begin{minipage}[t]{0.49\linewidth}
 \centering
   {\includegraphics[width=0.99\textwidth]{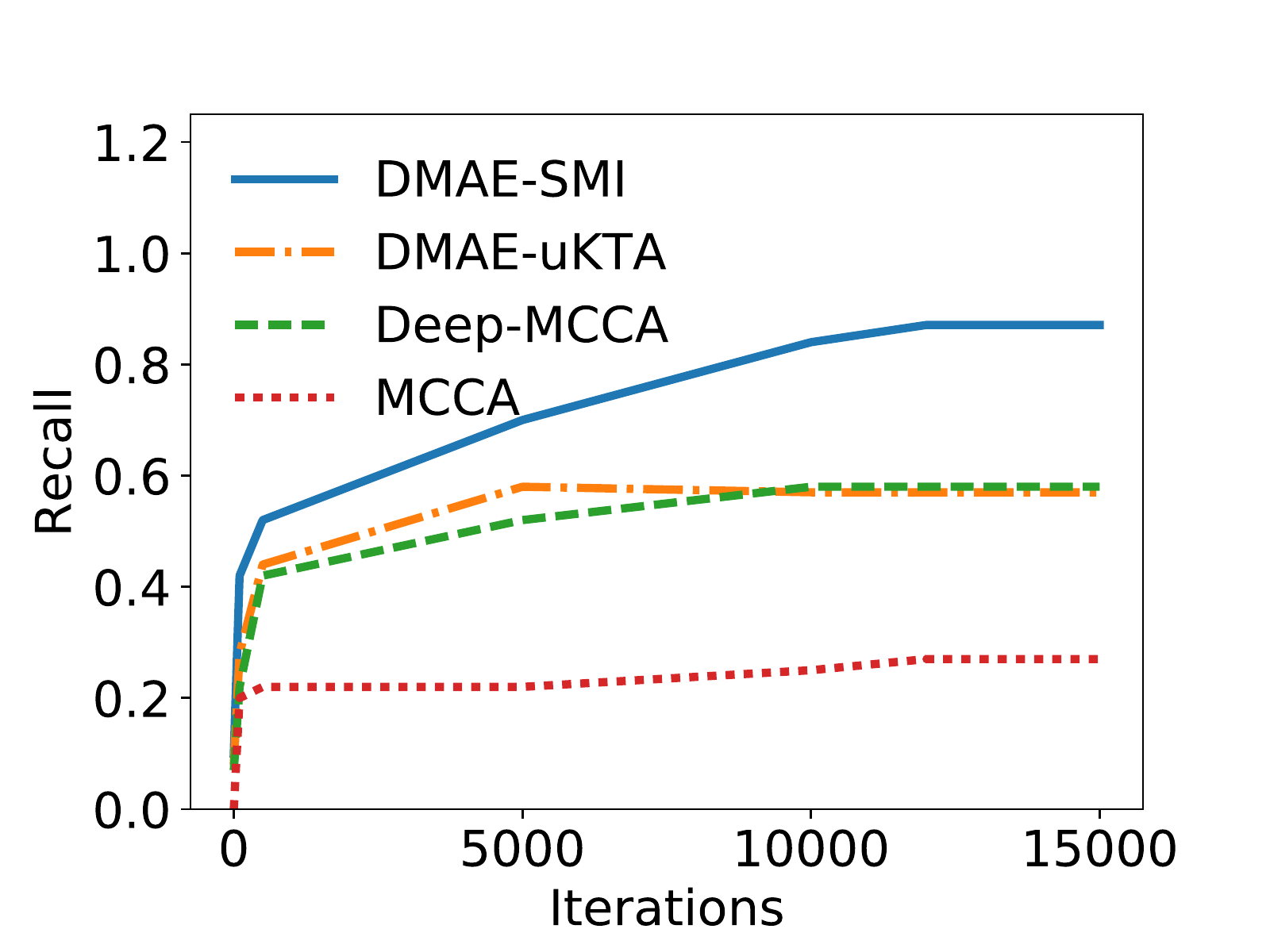}} \\ %\vspace{-0.10cm}
   (b) Recall. 
 \end{minipage}
 \centering
 \caption{Evolution of $\boldPi$ (label) matrix prediction accuracy during unsupervised classifier learning.}% (a) Precision. (b) Recall.}
 \label{fig:precision_recall_pi}
 \end{center}
 \end{figure}

To see how the accuracy of $\boldPi$ estimation changes during learning, we visualise the mean precision and recall over learning iterations in Figure~\ref{fig:precision_recall_pi}. We can see that: (i) Precision and recall rise monotonically over time before eventually asymptoting. (ii) DMAE-SMI performance grows faster and converges to a higher point than the alternatives.

\keypoint{Results: Testing Accuracy} To complete the evaluation of the actual learned classifier, we next assume that the estimated $\widehat{\boldPi}$ label matrix is correct, and use these labels to train a SVM classifier, which is evaluated on the testing split of each dataset.  The results for CIFAR-10 and AwA are shown in Tables~\ref{tab:SSLUnsupervised classifierCIFAR} and \ref{tab:SSLUnsupervised classifierAWA} respectively. As shown by the L/U/T proportions listed, supervised (all training data pairs given), semi-supervised (some training pairs given), semi-supervised transductive, and unsupervised (no training pairs given) settings are evaluated. From the results we can see that: (i) Our unsupervised learned classifiers work well (eg DMAE-SMI in 0-40-60 condition on AwA is 32\% vs 2\% chance). (ii) Our method can exploit semi-supervised learning (eg in CIFAR the 10-40-10 vs 10-0-10 condition shows the benefit of exploiting 80\% unlabeled train data in addition to 20\% labeled train data). (iii) We can also exploit seamlessly semi-supervised transductive learning if unpaired testing data is available (top vanilla vs bottom transductive condition).  (iv) Our method outperforms alternatives in each experiment (DMAE-SMI vs others).

\begin{table}[t]
\centering
\caption{Classification accuracy on CIFAR-10 test set. The data is split into labeled training data (L), unlabeled training data (U) and test data (T) in units of 10,000. Train on L and U, test on T.} \label{tab:SSLUnsupervised classifierCIFAR}
\resizebox{.5\textwidth}{!}{
\begin{tabular}{|lcccc|}
\hline 
\multicolumn{5}{|c|}{CIFAR-10}\tabularnewline
 \hline 
 L-U-T data (10K) & DMAE-SMI & DMAE-uKTA  & Deep-MCCA & MCCA \tabularnewline
\hline
~0-50-10 & {\bf 0.70} & 0.59 & 0.20 & 0.15 \\ 
10-40-10 & {\bf 0.86} & 0.73 & 0.40 & 0.26 \\
10-~0-10 & {\bf 0.72} & 0.62 & 0.21 & 0.17  \\
50-~0-10 & {\bf 0.91} & 0.84 & 0.57 & 0.52 \\ 
\hline
\hline
\multicolumn{5}{|c|}{CIFAR-10 Transductive}\tabularnewline
\hline
~0-50-10 & {\bf 0.76} & 0.66 & 0.22 & 0.16 \\ 
10-40-10 & {\bf 0.89} & 0.80 & 0.45 & 0.29 \\
10-~0-10 & {\bf 0.75} & 0.64 & 0.23 & 0.16 \\
50-~0-10 & {\bf 0.92} & 0.85 & 0.60 & 0.54 \\ 
\hline
\end{tabular}}
\end{table}

\begin{table}[t]
\centering
\caption{Classification accuracy on AWA test set (50-way).  The data is split into labeled training data (L), unlabeled training data (U) and test data (T) with listed \%. Train on L and U, test on T.} \label{tab:SSLUnsupervised classifierAWA}
\resizebox{.5\textwidth}{!}{
\begin{tabular}{|lcccc|}
\hline 
\multicolumn{5}{|c|}{AwA}\tabularnewline
 \hline 
L-U-T data (\%)  & DMAE-SMI & DMAE-uKTA  & Deep-MCCA & MCCA \tabularnewline
 \hline 
~0-40-60 & {\bf 0.32} & 0.20 & 0.16 & 0.12 \tabularnewline
20-~0-60 & {\bf 0.40} & 0.28 & 0.15 & 0.11 \tabularnewline
20-20-60 & {\bf 0.47} & 0.32 & 0.18 & 0.14 \tabularnewline
40-~0-60 & {\bf 0.60} & 0.48 & 0.24 & 0.18 \tabularnewline 
\hline
\hline
\multicolumn{5}{|c|}{AWA Transductive}\tabularnewline
\hline
~0-40-60 & {\bf 0.42} & 0.28 & 0.21 & 0.16 \tabularnewline
20-20-60 & {\bf 0.54} & 0.40 & 0.22 & 0.17 \tabularnewline
20-~0-60 & {\bf 0.52} & 0.38 & 0.22 & 0.17 \tabularnewline
40-~0-60 & {\bf 0.70} & 0.52 & 0.28 & 0.20 \tabularnewline 
\hline
\end{tabular}}
\end{table}

\section{Conclusion}
\label{sec:Conclusion}
We proposed \emph{\mname} (\mabbr~), which are capable of learning pairing and a common latent space from \emph{unpaired} multi-modal. The ~\mabbr~~ approach elegantly spans unsupervised, semi-supervised, fully supervised and zero-shot settings. In the supervised setting it is competitive with state of the art alternatives on captioning. In the less studied semi-supervised and unsupervised settings it outperforms the few existing alternatives. Prior studies of unsupervised cross-domain matching have generally used toy data. We showed that our \mabbr ~~approach can scale to real vision and language data and solve a novel unsupervised classifier learning problem. Although we evaluated our approach on captioning and classifier learning, it is a widely applicable cross-modal method with potential applications from person re-identification \cite{feng2014correspondenceAutoencoder} to word-alignment \cite{DBLP:journals/corr/abs-1710-04087}, which we will explore in future work.

\clearpage
\small
\bibliographystyle{ieee}
\bibliography{main}

\end{document}

%% file: mathdef.tex
\newtheorem{defi}{Definition}

\newtheorem{lemm}[defi]{Lemma}

\newcommand{\argmin}{\mathop{\mathrm{argmin\,}}}
\newcommand{\argmax}{\mathop{\mathrm{argmax\,}}}

\newcommand{\diag}[1]{\mathrm{diag}\left(#1\right)}

\newcommand{\mathbbR}{\mathbb{R}}

\newcommand{\boldone}{{\boldsymbol{1}}}

\newcommand{\boldA}{{\boldsymbol{A}}}

\newcommand{\boldH}{{\boldsymbol{H}}}
\newcommand{\boldI}{{\boldsymbol{I}}}

\newcommand{\boldK}{{\boldsymbol{K}}}
\newcommand{\boldL}{{\boldsymbol{L}}}

\newcommand{\boldf}{{\boldsymbol{f}}}
\newcommand{\boldg}{{\boldsymbol{g}}}
\newcommand{\boldh}{{\boldsymbol{h}}}

\newcommand{\boldx}{{\boldsymbol{x}}}
\newcommand{\boldy}{{\boldsymbol{y}}}

\newcommand{\boldalpha}{{\boldsymbol{\alpha}}}

\newcommand{\boldTheta}{{\boldsymbol{\Theta}}}

\newcommand{\boldPi}{{\boldsymbol{\Pi}}}